\documentclass{article}

\usepackage{microtype}
\usepackage{graphicx}
\usepackage{subfigure}
\usepackage{booktabs} 
\usepackage{amsfonts}       
\usepackage{amsmath}
\usepackage{mathtools}

\usepackage{hyperref}

\usepackage[accepted]{tccmlicml2021}

\icmltitlerunning{EVGen: Adversarial Networks for Learning Electric Vehicle Charging Loads and Hidden Representations}

\begin{document}

\setlength{\abovedisplayskip}{2pt}
\setlength{\belowdisplayskip}{2pt}

\twocolumn[
\icmltitle{EVGen: Adversarial Networks for Learning \\ 
Electric Vehicle Charging Loads and Hidden Representations}



\icmlsetsymbol{equal}{*}

\begin{icmlauthorlist}
\icmlauthor{Robert Buechler}{equal,stanME}
\icmlauthor{Emmanuel Balogun}{equal,stanME}
\icmlauthor{Arun Majumdar}{stanME}
\icmlauthor{Ram Rajagopal}{stanCEE}

\end{icmlauthorlist}

\icmlaffiliation{stanME}{Department of Mechanical Engineering, Stanford University, Stanford, California, United States}
\icmlaffiliation{stanCEE}{Department of Civil and Environmental Engineering, Stanford University, Stanford, California, United States}

\icmlcorrespondingauthor{Robert Buechler}{rbuec@stanford.edu}

\icmlkeywords{Machine Learning, ICML}

\vskip 0.3in
]



\printAffiliationsAndNotice{\icmlEqualContribution} 

\begin{abstract}
    
    The nexus between transportation, the power grid, and consumer behavior is more pronounced than ever before as the race to decarbonize the transportation sector intensifies. Electrification in the transportation sector has led to technology shifts and rapid deployment of electric vehicles (EVs). The potential increase in stochastic and spatially heterogeneous charging load presents a unique challenge that is not well studied, and will have significant impacts on grid operations, emissions, and system reliability if not managed effectively. Realistic scenario generators can help operators prepare, and machine learning can be leveraged to this end. In this work, we develop generative adversarial networks (GANs) to learn distributions of electric vehicle (EV) charging sessions and disentangled representations. We show that this model structure successfully parameterizes unlabeled temporal and power patterns without supervision and is able to generate synthetic data conditioned on these parameters. We benchmark the generation capability of this model with Gaussian Mixture Models (GMMs), and empirically show that our proposed model framework is better at capturing charging distributions and temporal dynamics.
    
\end{abstract}

\section{Introduction}

    The United States, China, and the European Union collectively emit over half of the global greenhouse gas emissions \cite{epa_2021}, highlighting their critical role in combating global climate change. The transportation sector accounts for 29\% of total Greenhouse Gas (GHG) emissions in the U.S, and 14\% globally \cite{epa_2021}, making the decarbonization of the transportation sector critical to environmental sustainability. The resurgent urgency to curtail emissions, driven by physical manifestations of climate change today, has motivated governments to create policy that is driving massive Electric Vehicle (EV) adoption. The share of electricity consumption on the power grid attributable to charging EVs is expected to increase up to 40-fold in the United States between 2019 and 2030 \cite{IEA_2020}. EVs accounted for 54\% of new car sales in Norway in 2020, up from 42\% in 2019 \cite{reuters_2021}. With the 2035 net-zero carbon goal in the U.S and other electrification targets set by China, UK, and others, we can expect a dense influx of EVs on the roads over the next decade. However, with increased EV penetration, the subsequent massive increase in stochastic power and energy demand from the grid will accelerate the aging of grid assets, complicate optimal operation of power systems, and could require massive upgrades of power systems components. Understanding possible future states of the grid via scenario simulation is paramount to effectively electrifying transportation at scale.
    
    Some work has been done on developing statistical representations of EV charging using Gaussian Mixture Models (GMM) from real charging data \cite{8442988} \cite{POWELL2020115352}. Others have leveraged travel survey data to estimate future EV charging from current driver travel patterns \cite{7460088} using traditional statistical approaches. However, few studies utilize deep learning (DL) based approaches, like generative adversarial networks (GAN) \cite{Goodfellow_Pouget-Abadie_Mirza_Xu_Warde-Farley_Ozair_Courville_Bengio_2014}.  
    
    GANs were originally developed for single and multi-channel imagery, and GAN architectures continue to be optimized for this application. GANs have also been studied for learning 1D data distributions, including financial timeseries datasets \cite{De_Meer_Pardo} and medical data anonymization \cite{Bae_Jung_Yoon_2019}\cite{Yoon_Drumright_van_der_Schaar_2020}. Multiple techniques have been developed for conditioning GANs on inputs. A supervised example of this is CGAN \cite{Mirza_Osindero_2014}. InfoGAN \cite{Chen_Duan_Houthooft_Schulman_Sutskever_Abbeel_2016} proposes maximizing the mutual information between latent variables and the output by using a third network, while SCGAN \cite{Li_Chen_Wang_Wu_Tong_2019} computes a similarity constraint inside the generator network. As InfoGAN and SCGAN are both unsupervised frameworks, they can be used to learn disentangled representations without labeled data. 

\section{Dataset}

    The dataset used in this work is from a real EV charging network at a large tech company campus. The raw data comprises roughly 20000 individual charging events, including variables such as the driver ID, plug-in event ID, charging start and end time, average interval power, energy, driver zip code, and others which are not critical for this paper. Each charging event records power and energy in 15-minute intervals during the charging event. We standardize the length of all data samples into a 24 hour period in 15-minute intervals and assign the charging interval window to a predefined standard window, padding the sequence at intervals where each plug-in event has no charging powers.  Thus, one data-sample is a 96 X 1 vector representing a charge event. The uniform sequence length simplifies the model and makes it natural to capture representations attributable to charging behavior, battery types, etc. The data is then split between training and test sets with a 95:5 ratio, and is then normalized for training. 

\section{Experiment}

    In this work, we utilize a GAN to generate synthetic charging load curves in a specific charging location, conditioned on the learned disentangled representations. We also generate synthetic load curves using a baseline model (GMM) and compare the performance.
    
\subsection{Model Structure}

    \begin{table*}[t]
    \caption{Properties of the CNN-based network. N is the size of the minibatch.}
    \begin{center}
    \begin{footnotesize}
    \begin{tabular}{ |c|c|l| } 
    \hline
    \multicolumn{3}{|c|}{Generator} \\
    \hline
    Layer Type & Parameters & Output Shape \\ 
    \hline
    Input (continuous case) & $[z,c]\in\mathbb{R}^{88}$, where $z\sim unif(0,1)\in\mathbb{R}^{80}$ and $c\sim unif(0,1)\in\mathbb{R}^{8}$ & (N, 88)  \\
    Input (discrete case) & $[z,c]\in\mathbb{R}^{88}$, where $z\sim unif(0,1)\in\mathbb{R}^{80}$ and $c=e_{i}\in\mathbb{R}^{8}$ & (N, 88)  \\
    Fully Connected & Units=100, LeakyReLU(0.2) & (N, 150) \\
    Expand Dimension & -- & (N, 1, 150) \\
    Conv1D & Filters = 32, kernel size= 5, padding =’replicate’, LeakyReLu(0.2) & (N, 32,150) \\
    Conv1D & Filters = 16, kernel size= 5, padding =’replicate’, LeakyReLu(0.2) & (N, 16, 150) \\
    Conv1D & Filters = 8, kernel size= 5, padding =’replicate’, LeakyReLu(0.2) & (N, 8, 150) \\
    Conv1D & Filters = 1, kernel size= 5, padding =’replicate’, LeakyReLu(0.2) & (N, 1, 150) \\
    Squeeze Dimension & -- & (N, 150) \\
    Fully Connected & Units=125, LeakyReLU(0.2) & (N, 125) \\
    Fully Connected & Units=100, LeakyReLU(0.2) & (N, 100) \\
    Fully Connected & Units=96, LeakyReLU(0.2) & (N, 96) \\
    \hline
    \hline
    \multicolumn{3}{|c|}{Discriminator / Critic} \\
    \hline
    Layer Type & Parameters & Output Shape \\ 
    \hline
    
    Input & -- & (N, 96)  \\ 
    Expand Dimension & -- & (N, 1, 96) \\
    Conv1D & Filters = 32, kernel size= 5, padding =’replicate’, LeakyReLu(0.2) & (N, 32, 96) \\
    MaxPool & PoolSize = 2 & (N, 32, 48) \\
    Conv1D & Filters = 16, kernel size= 5, padding =’replicate’, LeakyReLu(0.2) & (N, 16, 48) \\
    MaxPool & PoolSize = 2 & (N, 16, 24) \\
    Conv1D & Filters = 8, kernel size= 5, padding =’replicate’, LeakyReLu(0.2) & (N, 8, 24) \\
    Flatten & -- & (N, 192) \\
    Fully Connected & Units=50, LeakyReLU(0.2) & (N, 50) \\
    Fully Connected & Units=15, LeakyReLU(0.2) & (N, 15) \\
    Fully Connected & Units=1, LeakyReLU(0.2) & (N, 1) \\
    \hline
    \end{tabular}
    
    \label{table:CNN model specs}
    \end{footnotesize}
    \end{center}
    \end{table*}

    We utilize a GAN variant to learn the probability distribution of the data. Specifically, we use a GAN with Wasserstein loss (WGAN) \cite{Arjovsky_Chintala_Bottou_2017} with a gradient penalty to satisfy the Lipschitz constraint (WGAN-GP) \cite{Gulrajani_Ahmed_Arjovsky_Dumoulin_Courville_2017}. The gradient penalty term replaces the need for weight clipping, a technique that can limit the models learning ability and lead to undesired behaviour. Wasserstein loss has been shown to improve the stability of training and address mode collapse. We use a combination of stacked convolutional layers and fully connected layers in both the generator (G) and disciminator (D)\footnote{This network is also commonly called a critic, since the network is simply trying to maximize the output of real instances, rather than classify between and fake. We will continue using "discriminator" terminology, although "critic" is also accurate.} networks. The specific architecture for each network is shown in Table \ref{table:CNN model specs}. 
    
    We extend the WGAN-GP architecture to learn disentangled representations. Representation learning is important for many tasks and is popularly applied in object detection and image recognition. It is beneficial to be able to preform this task on unlabeled data, since it more difficult to acquire large open-sourced labeled EV charging datasets. This allows unsupervised algorithms to be trained on anonymized or poorly labeled data. For the unsupervised learning of disentangled representations, we utilize a term (Equation \ref{eq:SC}) from part of the SCGAN framework. This term encourages the generator to learn the difference between samples of different types and is calculated pair-wise on the training set. This is a simpler method than InfoGAN, since SCGAN shares parameters with G and does not require training a third network. This learning can be done with continuous variables to learn attributes that change smoothly, or discrete variables to denote the boundaries between clearly-defined clusters. The similarity constraint (SC) between condition $c$ and synthetic generated data for continuous conditional variables is written as
    \begin{multline}
    SC(\bold{x},c) = \frac{1}{N(N-1)} \sum_{i}\sum_{j\neq i} \\
    \left((1-|c_i-c_j|)sim(\bold{x}_i,\bold{x}_j)+\frac{|c_i-c_j|}{sim(\bold{x}_i,\bold{x}_j)}\right)
    \label{eq:SC}
    \end{multline}
    
    where $N$ is the size of the minibatch, $x$ is synthetic data from the generator, and $i,j \in [1, N]$. We take $sim(\bold{x}_i,\bold{x}_j)=\|\bold{x}_i-\bold{x}_j\|_2$. This term is minimized by the G network and is regularized with $\lambda_{SC}$. During training, each $c$ is drawn from $unif(0,1)$. For discrete variables, the $(1-|c_i-c_j|)$ term in Eqn \ref{eq:SC} is replaced with $<\bold{c}_i, \bold{c}_j>$ and $|c_i-c_j|$ (numerator) is replaced with $(1 - <\bold{c}_i, \bold{c}_j>)$, where $\bold{c}$ is a one-hot vector sampled from a multi-nomial distribution. The combination of the SC calculation with WGAN-GP yields a model that we refer to as SC-WGAN-GP. We use the ADAM optimizer, an adaptive learning rate scheduler, and mini-batches with 256 samples per batch. 

\subsection{Benchmark Model}
    A Gaussian Mixture Model (GMM) is a convex combination of Gaussians. Each Gaussian represents a group or cluster, and each cluster k, has its own weight $\alpha_{k}$, mean $\mu_{k}$, and variance $\sigma^{2}_{k}$, with the mean and variances parametrizing the Gaussian density function for respective clusters. The GMM in its most general form is expressed for a multidimensional variable, where $\mu_{k}$ is a vector and $\Sigma_{k}$ is the covariance matrix. The weights $\alpha_{k}$, represent probabilities and must sum to one.
        \begin{equation}
            \sum_{k=1}^{K} \alpha_{k} = 1
        \end{equation}
    
    The parameters for the GMM are obtained via an Expectation Maximization (EM) Algorithm, which seeks to find parameters that maximize the likelihood of the observed data, given the distribution defined by the parameters. If we let our parameters be tuple $\bold{\theta }= (\alpha, \mu, \Sigma)$. The algorithm solves the optimization problem to maximize the likelihood,
     \begin{equation}
        \begin{multlined}
        \bold{Q(\bold{\theta^{*}, \theta})} = \sum_{N=1}^{N}\sum_{k=1}^{K} \gamma(z_{nk})[\ln{\alpha_{k}} + \\ \ln\mathcal{N}(x_{n}|\mu_{k}, \Sigma_{k})] - \lambda(\sum_{k=1}^{K}\alpha_{k} - 1)
        \end{multlined}
    \end{equation}
    given the observed data true distribution. $\gamma$ represents the probability of a particular cluster k, given the observed data. The last term includes the Lagrange multiplier $\lambda$, which serves to constrain the sum of the Gaussian weights to one. 
    To develop the GMM, we postulate that a simplified EV charging session is completely described by charging session start time, charging session duration, and charging session average load intensity. Thus, the GMM models the prior and the joint probability given each prior. This means it learns the probability for each cluster and captures the joint probability of charge start times, charging duration, and load intensities of charge sessions, given some cluster k with weight (prior probability) $\alpha_{k}$. GMMs use the popular Expectation Maximization (EM) algorithm to iteratively solve for a minimum.

\subsection{Results}

    \begin{figure*}[t]
        \centering
        \includegraphics[width=6.75in, trim={0 68cm 0 0}, clip=true]{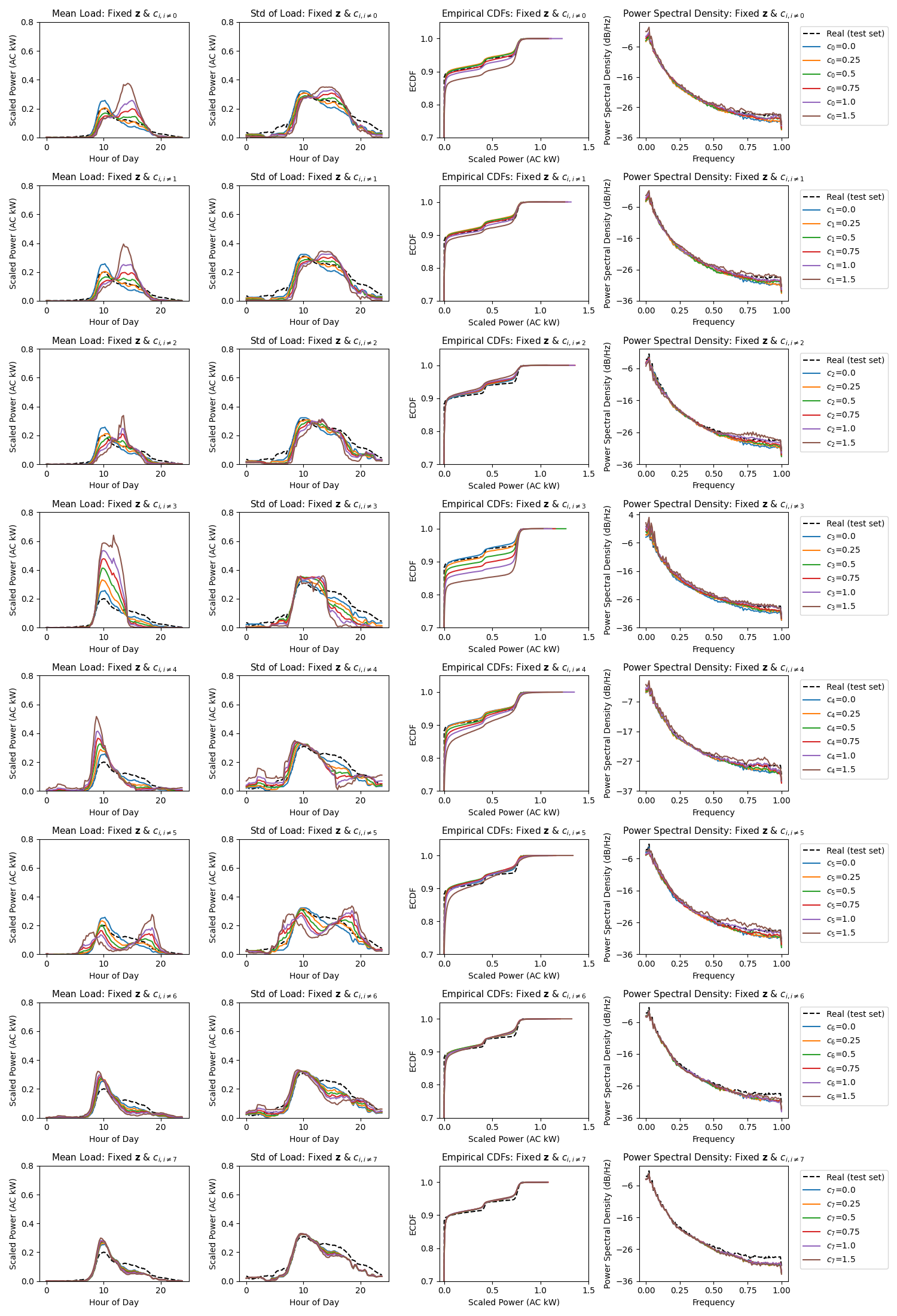}
        \includegraphics[width=6.75in, trim={0 19.5cm 0 48.5cm}, clip=true]{Figures/SCGAN_clusters.png}
        \caption{Statistics of SC-WGAN-GP (continuous) generator output. Only two continuous variables $c_0$ (top) and $c_5$ (bottom) are shown. Each line represents a different continuous variable value.}
        \label{fig:gan clusters}
    \end{figure*}
    
    \begin{figure*}[!h]
        \centering
        \includegraphics[width=6.75in]{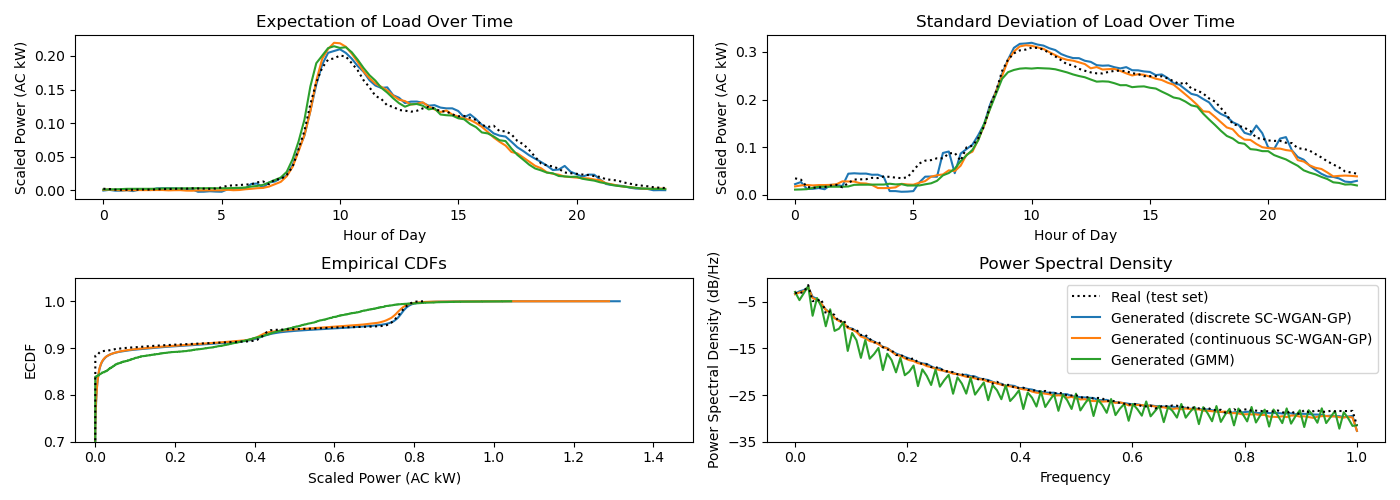}
        \caption{Summary statistics for the three networks: continuous SC-WGAN-GP, discrete SC-WGAN-GP, and GMM. The GAN architectures most notably outpreform the GMM when considering the the CDF and PSD of the signals.}
        \label{fig:model results}
    \end{figure*}
    
    We train our benchmark (the GMM) and SC-WGAN-GP models and compare performance. For this experiment, we choose a generator with an 80-dimensional input latent space $z$ from $unif(0,1)$. For representation learning, we train one model with eight continuous variables and another model with eight discrete variables. The SC-WGAN-GP trains for around 4000 epochs and takes about 9hrs to converge on a NVIDIA K80 GPU, although it can produce good (visually, qualitative) fake samples after around 500 epochs. The training curves for both networks can be viewed \href{https://tensorboard.dev/experiment/ngVr9qJVQUG1Rg4QLX7lrg/}{\emph{here}} for the continuous SC-WGAN-GP and \href{https://tensorboard.dev/experiment/9vnEY101Q0mTDJqBwQTxPg/}{\emph{here}} for the discrete SC-WGAN-GP. The D network, SC, and GP terms are stable and converge. 
    
    For the GMM, we selected the cluster count to be 1000, as this was the main parameter we tuned for this application. The maximum number of iterations for the Expectation Maximization (EM) algorithm was set to 50000 and the convergence tolerance was $10^{-6}$. The GMM displays improving performance with the number of clusters, up to a limit, at which it starts to overfit. We find the optimal cluster count N to be around the chosen value of 1000.  The GMM converges in about 6 - 10 minutes, which is much faster than the SC-WGAN-GP. Performance statistics for various numbers of clusters are shown in Figure A1 in the supplementary materials.

\subsection{Discussion}

    It is imperative that the generated data can display mode diversity (the ability of the model to capture multiple modes from the training set, not just one) and statistical resemblance (the generated output should have similar statistical properties as the real distribution). It is insignificant to compare single samples from each distribution; the entire distributions must be analyzed as a whole. To analyze the probability of certain-valued loads being present in each distribution, the cumulative distribution function (CDF) is derived empirically from the real and generated sample loads. To analyze the periodic component and the temporal fluctuations present in each distribution, the power spectral density is calculated, as is shown along with the CDF in Figure \ref{fig:model results}.

    For the GAN, we observe a trade-off between generator performance and representation learning. If $\lambda_{SC}$ is high, the generator may not converge, and if $\lambda_{SC}$ is low, the generator may not learn to capture representations. $\lambda_{SC}$ therefore requires careful tuning, and may benefit from dynamic scheduling; although we saw no immediate successes of such a structure in our experiments, it should be explored further. The authors of SCGAN implement the pair-wise loss calculation as a nested iteration loop for each minibatch parameter update step, and cite their method as being 3-5x slower than InfoGAN. We implement a fully vectorized version of the SC pair-wise calculation that is at least 50x faster than SCGAN. 
    
    Figure \ref{fig:gan clusters} shows two of the eight continuous variables learned by SC-WGAN-GP. $c_0$ appears to increase the power level of charging sessions in the afternoon in exchange of decreasing sessions in the morning with higher values of $c_0$. $c_5$ appears to shift consumption out to later in the evening while drastically reducing midday consumption at higher $c_5$ values. We found that other variables (not shown) learned various other temporal and power-level patterns. The value for 1.5 for each variable demonstrates model extrapolation since values were sampled from $unif(0,1)$ during training. This is significant for this application because it offers the capability to test future scenarios where certain variables are expected to change, on large unlabeled datasets. 
    
    Figure A2 in the supplementary materials show the discrete representations learned by SC-WGAN-GP. It is clear that the discrete variables are grouping load shapes by time and charging duration.
    
    An inherent limitation of the GMM model is its inability to capture transient charging dynamics for each generated charging session, since the model assumes an average steady load for each session. One possible mitigation for this will be to sample from one or very few similar GMM clusters multiple times, and then take the average for each time-interval. With this idea, the GMM still fails to produce realistic loads but instead, produces stepped load shapes. Because the GMM model is fast and performs okay, it is better suited for long-term planning of power infrastructure, but not real-time operation and simulation. For operations and control, transient dynamics are important as the power system and all components that connect to it are predominantly dynamic. Additionally, because the GMM models steady loads, information on the frequency driven properties that might aid the learning of subtle and latent representative groups like the EV car type, model, or battery chemistry, are lost.

\section{Conclusion}

    In this work, we built a framework that successfully captures hidden representations in an EV charging dataset in an unsupervised manner. The results show that our proposed framework, SC-WGAN-GP, is able to generate statistically realistic EV load curves, and can learn hidden temporal and power patterns in the training set better than more traditional modeling methods. Also importantly, the latent variables are intuitive and have physical meaning; for instance, some variables group drivers with vehicles that charge fast, while some other variables capture temporal charging behavior from the distribution. This shows promising signs of developing representative EV charging scenarios for regions globally, which will be instrumental for grid infrastructural planning/operation. DL-based methods like GANs  could be used for EV load generation for demand response or vehicle-to-grid applications, where accurate short-term (second to minute scale) representations of load behaviour is critical. We propose the following future improvements to the model architecture and training procedure:
    
    \begin{itemize}
    
        \item \emph{Contrained SC loss.} Some continuous variables learned very similar patterns during training, resulting in redundant learning. We aim to constrain SC loss to learn dissimilar embeddings to minimize statistical redundancy in variables. 
        
        \item \emph{More interpretable model extrapolation.} The SC-WGAN-GP (continuous) is able to extrapolate learned representations past the distribution seen in the training set. However, the form of this extrapolation is unknown, and may not be linear. Mapping these learned representations to a more interpretable space would help in real applications, i.e. increasing the value of a variable that has learned battery size.
        
        \item \emph{Regional Learning.} Although the data used in this work is limited to a single region, we can train our model to learn distributions for different regions by conditioning the GAN outputs on locational variables.
        
        \item \emph{Address bias in training data.} As our training data was limited to a single company campus location, we aim to test the model's generalizability by training with public EV charging station data.

    \end{itemize}

\section*{Acknowledgements}

    Thank you to Siobhan Powell and Gustavo Vianna Cezar for their advice and support, including suggestions related to benchmark models and data sources. Thank you to our reviewers for providing helpful insight. 



\bibliography{references}
\bibliographystyle{icml2021}

\setcounter{figure}{0}   
\makeatletter 
\renewcommand{\thefigure}{A\@arabic\c@figure}
\makeatother

\onecolumn

\newpage

\textbf{\large{Supplementary Materials}}

    \begin{figure*}[h]
        
        \centering
        \includegraphics[width=2.5cm]{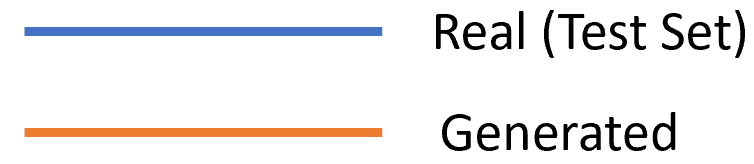}
        \includegraphics[width=6.75in]{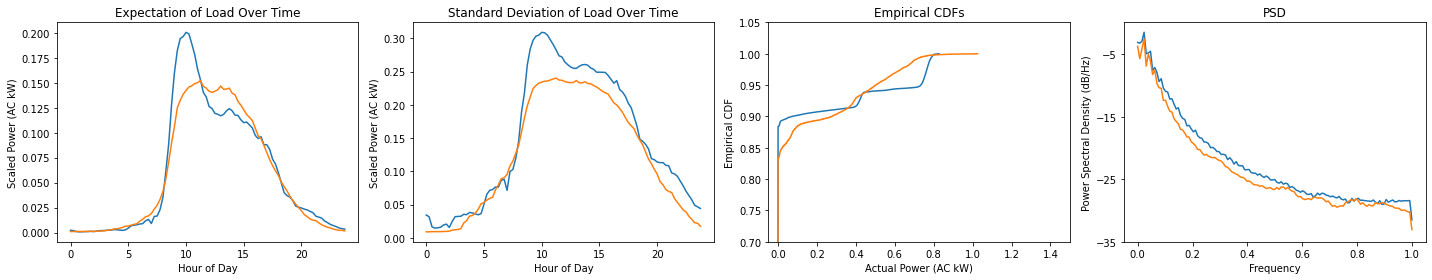}
        \includegraphics[width=6.75in]{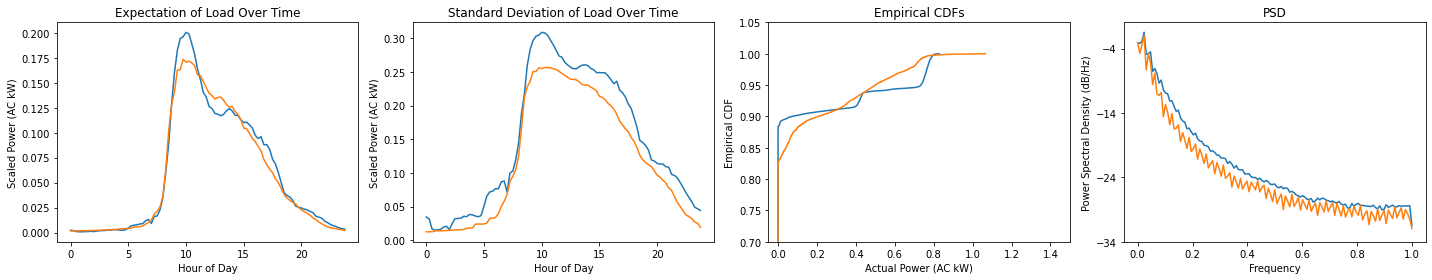}
        \includegraphics[width=6.75in]{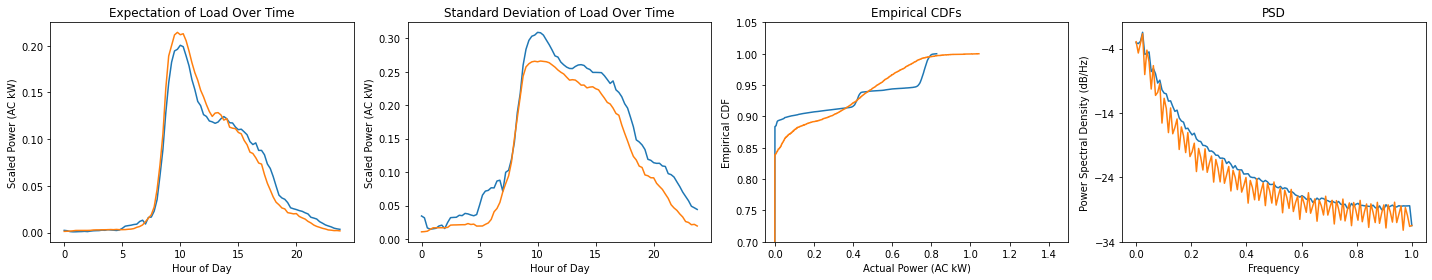}
        \includegraphics[width=6.75in]{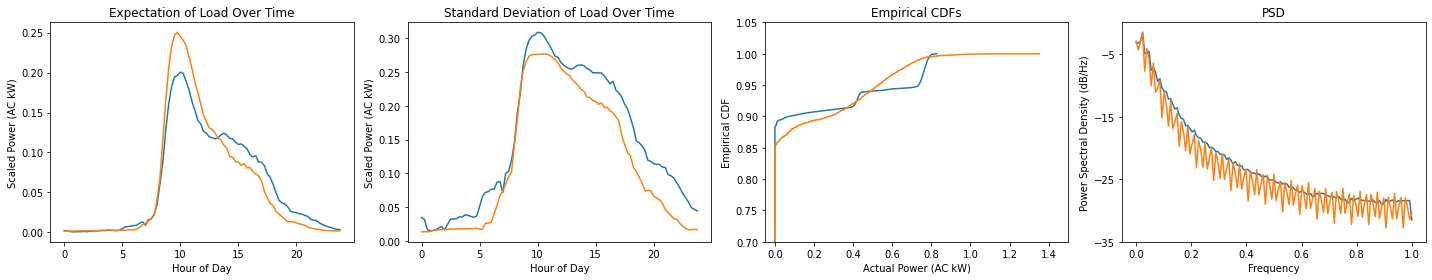}
        
        \caption{Summary statistics for GMM Number of clusters (N = 125, 500, 1000, 2000. Top to bottom). The GMM performance degrades for N = 2000.}
        \label{fig:GMM experiments results}
    \end{figure*}
    
    \begin{figure*}
        \centering
        \includegraphics[width=6in]{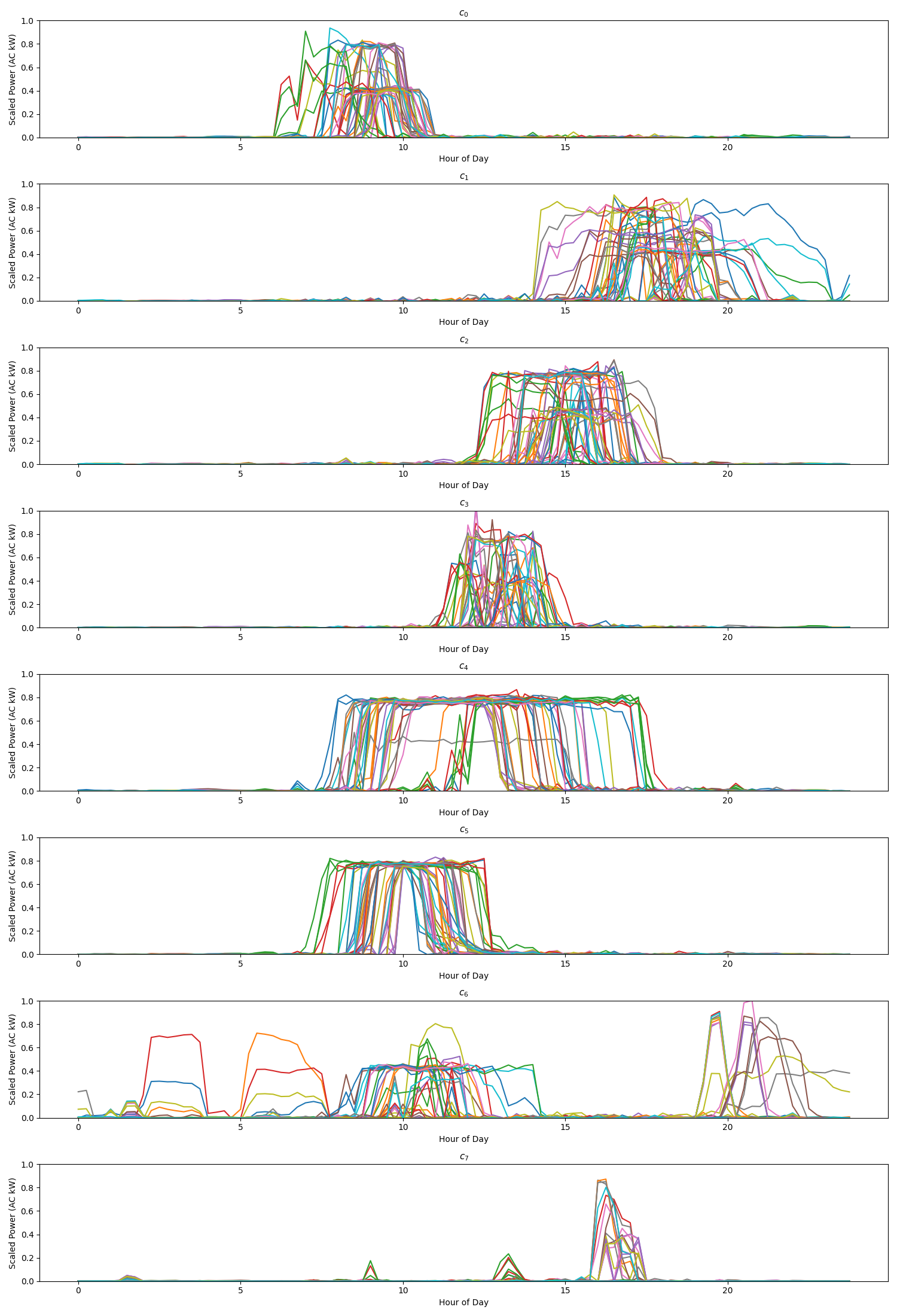}
        \caption{Sample synthetic data generated from the discrete SC-WGAN-GP (50 samples per plot). All plots were generated using the same latent space $z$. Each row includes samples from a different discrete variable.}
        \label{fig:discrete gan outputs}
    \end{figure*}

\newpage


        
        
    

\end{document}